\documentclass[letterpaper]{article} 
\usepackage{aaai23}  
\usepackage{times}  
\usepackage{helvet}  
\usepackage{courier}  
\usepackage[hyphens]{url}  
\usepackage{graphicx} 
\usepackage{natbib}  
\usepackage{caption} 
\usepackage{algorithm}
\usepackage{algorithmic}
\usepackage{array}
\usepackage{multirow}
\usepackage{multicol}
\usepackage{color}
\usepackage{caption}
\usepackage{subfigure}
\usepackage{amsfonts} 
\usepackage{booktabs}
\usepackage{amsmath}
\usepackage{amssymb}

\usepackage[normalem]{ulem}

\usepackage{newfloat}
\usepackage{listings}
\DeclareCaptionStyle{ruled}{labelfont=normalfont,labelsep=colon,strut=off} 
\floatstyle{ruled}
\newfloat{listing}{tb}{lst}{}
\floatname{listing}{Listing}
\title{FiTs: Fine-grained Two-stage Training for Knowledge-aware Question Answering}
\author{
    Bowen Cao\textsuperscript{\rm 1}\equalcontrib,Qichen Ye\textsuperscript{\rm 1}\equalcontrib, Nuo Chen\textsuperscript{\rm 3,4}, Weiyuan Xu\textsuperscript{\rm 1}, Yuexian Zou\textsuperscript{\rm 1,2}\thanks{Corresponding author.}
}
\affiliations{
    \textsuperscript{\rm 1}ADSPLAB, School of ECE, Peking University, Shenzhen, China,
    \textsuperscript{\rm 2}Peng Cheng Laboratory, Shenzhen, China\\
    \textsuperscript{\rm 3}Hong Kong University of Science and Technology (Guangzhou),
    \textsuperscript{\rm 4}Hong Kong University of Science and Technology

    \big\{yeeeqichen, zouyx\big\}@pku.edu.cn,
    \big\{cbw2021, xuwy\big\}@stu.pku.edu.cn,
    chennuo26@gmail.com
    
}

\usepackage{bibentry}

\begin{document}

\maketitle

\begin{abstract}
Knowledge-aware question answering (KAQA) requires the model to answer questions over a knowledge base, which is essential for both open-domain QA and domain-specific QA, especially when language models alone cannot provide all the knowledge needed.
Despite the promising result of recent KAQA systems which tend to integrate linguistic knowledge from pre-trained language models (PLM) and factual knowledge from knowledge graphs (KG) to answer complex questions, a bottleneck exists in effectively fusing the representations from PLMs and KGs because of \emph{(i)} the semantic and distributional gaps between them, and \emph{(ii)} the difficulties in joint reasoning over the provided knowledge from both modalities.
To address the above two problems, we propose a \textbf{Fi}ne-grained \textbf{T}wo-\textbf{s}tage training framework (FiTs) to boost the KAQA system performance: The first stage aims at aligning representations from the PLM and the KG, thus bridging the modality gaps between them, named knowledge adaptive post-training. The second stage, called knowledge-aware fine-tuning, aims to improve the model's joint reasoning ability based on the aligned representations.
In detail,  we fine-tune the post-trained model via two auxiliary self-supervised tasks in addition to the QA supervision.
Extensive experiments demonstrate that our approach achieves state-of-the-art performance on three benchmarks in the commonsense reasoning (\emph{i.e.}, CommonsenseQA, OpenbookQA) and medical question answering (\emph{i.e.}, MedQA-USMILE) domains.
\end{abstract}

\section{Introduction}

\begin{figure}[t]
\centering
\includegraphics[width=0.47\textwidth]{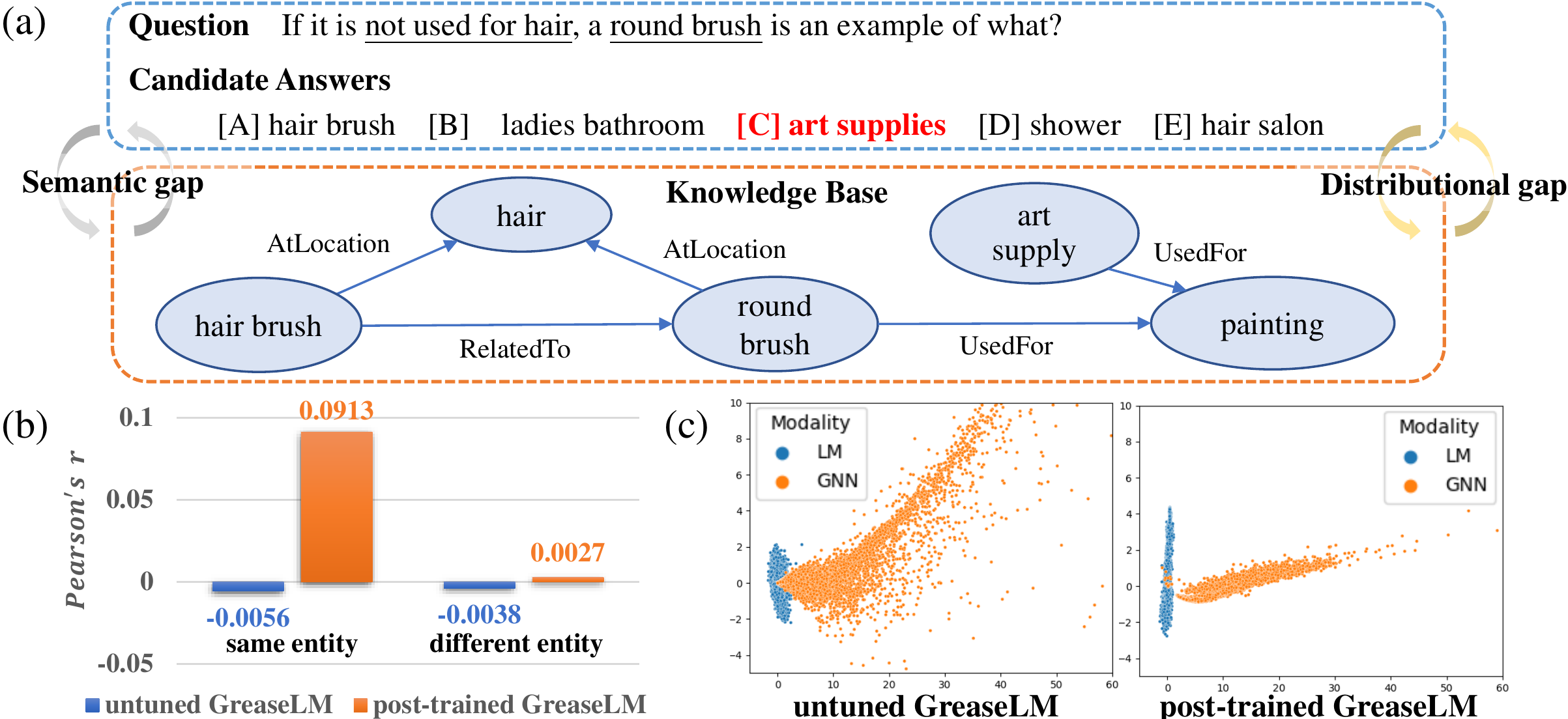} 
\caption{(a) An example of the knowledge-aware QA task from CommonsenseQA. (b) The Pearson correlation coefficient of representations
for the same/different entity 
from LM and GNN, which represents the semantic consistency between the two modalities (+1: positive correlation; -1: negative correlation; 0: no linear dependency). (c) PCA visualization of the distributions of entity representations from the untuned GreaseLM (integrating RoBERTa-Large and GAT), and the corresponding post-trained GreaseLM. The latter serves as a better starting point for fine-tuning.}
\label{Fig.example}
\end{figure}

Recent advances in large pre-trained language models (PLM) have demonstrated distinguishable applicability in various tasks. \cite{chen2021adaptive,M3ST, jin2022expectationmaximization, cao2022locvtp, TCNet}. 
However, empirical studies show that PLMs may struggle when dealing with examples that are distributionally different from the pre-training and fine-tuning corpora \cite{kassner2020negated}.
This shortcoming limits their performance in the open domain question answering (QA) task that requires a wide range of factual knowledge. More recently, some works \cite{obqaSplit, MHGRN} focus on the knowledge-aware question answering (KAQA) task (cf. Figure \ref{Fig.example}(a)), which allows access to external knowledge bases, especially knowledge graphs (KG), because KGs capture a broad coverage of factual knowledge explicitly using triplets that encode the relationships between entities. Taking advantage of both PLMs and KGs, these systems achieve remarkable results for tasks requiring open domain knowledge and structured reasoning \cite{kagneg19,MHGRN,QA-GNN}.

The overwhelming majority of state-of-the-art KAQA methods can be classified into two categories: semantic parsing-based (\emph{SP-based}) methods \cite{luo2018SP-based-1,sun2020SP-based-3} and information retrieval-based (\emph{IR-based}) methods \cite{chen2019IR-1,GreaseLM}. This paper mainly focuses on the latter which
follows a two-stage procedure: \emph{(i)} retrieving relevant knowledge from KGs under the information conveyed in the question; and then \emph{(ii)} fusing the retrieved knowledge and the contextualized representations captured by PLMs to perform joint reasoning. 

However, 
IR-based KAQA models inevitably suffer from two problems: (1) \textbf{Modality Gaps.}
There exist two intrinsic differences between two modalities, \emph{i.e.}, the semantic gap and distributional gap. On one hand, as shown in Figure \ref{Fig.example}(b), the weak dependency between representations of the same entity (\emph{e.g.},  ``\texttt{round brush}" in the QA context and in the knowledge base in Figure~\ref{Fig.example}(a)) from the untuned GreaseLM \cite{GreaseLM} suggests the \textbf{semantic gap} between the two modalities. On the other hand, we find that both the representations of the LM and the GNN are restricted to narrow cones with different shapes and apart from each other (which is consistent with the empirical findings of \citet{liang2022modality_gaps}), indicating the \textbf{distributional gap}; (2) \textbf{Difficulties in Joint Reasoning}, which cause the problem in training the model to answer questions over the provided two sources of knowledge. We further discuss it in section \textit{Knowledge-aware Fine-tuning}.

\begin{figure}[t]
\centering
\includegraphics[width=0.47\textwidth]{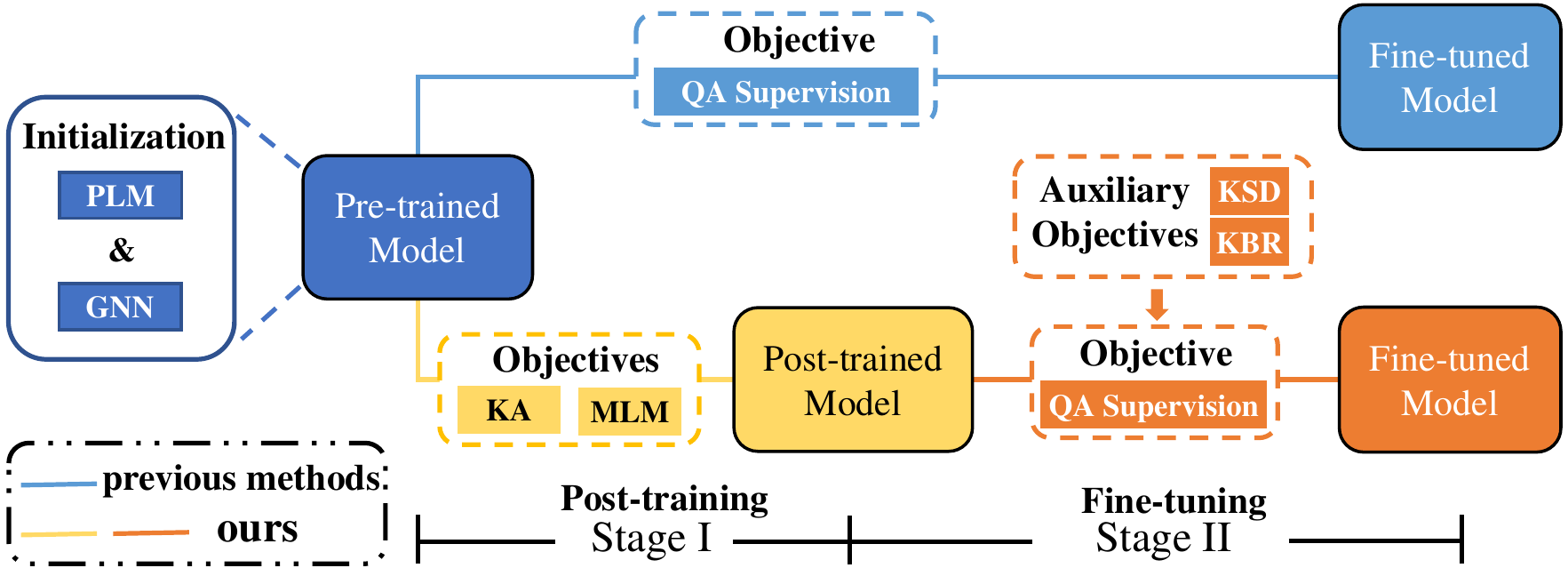} 
\caption{The pipelines of previous methods and our two-stage training framework for KAQA tasks.}
\label{Fig.pipeline}
\end{figure}

To address the above two problems, we propose a \textbf{Fi}ne-grained \textbf{T}wo-\textbf{s}tage training framework (\textbf{FiTs}), including post-training and fine-tuning stages (cf. Figure \ref{Fig.pipeline}). \textbf{In the first stage}, we present a simple but effective post-training method with the knowledge adaptive (\textbf{KA}) objective that aligns the representations from PLMs and KGs.
As shown in Figure \ref{Fig.example}(b), the semantic consistency between representations of the same entity is greatly improved after post-training; Figure \ref{Fig.example}(c) shows that the distributions of the two modalities are adapted to each other. They demonstrate that both the semantic and the distributional gaps are alleviated, leading to a better starting point for fine-tuning.
\textbf{In the second stage}, our motivation is to train the model to efficiently and effectively reason with both sources of knowledge. To this end, we develop two auxiliary self-supervised learning objectives—\emph{(i)} knowledge source distinction (\textbf{KSD}): we let the model distinguish whether a retrieved entity is related to the question, related to the answer, or an irrelevant entity to improve model's ability in knowledge understanding; and \emph{(ii)} knowledge backbone regularization (\textbf{KBR}): we impose a regularization term on the retrieved KG knowledge triplets to enhance the joint reasoning ability of the model—in addition to the supervision signal to perform fine-tuning.

Experimental results on three benchmarks (\emph{i.e.}, CommonsenseQA, OpenbookQA, and MedQA-USMILE) demonstrate that FiTs boost the model performance for multiple choice question answering, a typical task in KAQA. Specifically, our model achieves an absolute improvement in accuracy by 2.6\%, 1.8\%, and 0.6\% on the above three benchmarks, respectively, with only \textbf{1\%} additional trainable parameters, suggesting the effectiveness and the domain generality of the proposed method\footnote{The code can be found at \url{https://github.com/yeeeqichen/FiTs}}.

\section{Related Work}
Current KAQA methods can be categorized into two groups: IR-based approaches and SP-based approaches.

\textbf{SP-based} methods \cite{kapanipathi2020SP-based-2} reason over KGs with logic forms generated by conducting syntactic and semantic analysis on the question. Since the quality of the logic form is highly dependent on the parsing module which converts unstructured text into structured representations, complex questions with compositional semantics increase the difficulties in linguistic analysis, leading to a bottleneck in performance improvement. Meanwhile, manually annotating logic forms is costly and labor-intensive, which poses another limitation.

\textbf{IR-based} KAQA systems typically consist of the modules of KG retrieval and joint reasoning. For KG retrieval, we follow the procedure from \citet{QA-GNN} to retrieve relevant KG attributes.
To perform joint reasoning over PLMs and the retrieved KG sub-graphs, the most critical challenge is that models have to fuse the semantically and distributionally different knowledge encoded in PLMs and KGs.
Some works separately capture language representations and graph representations with two-tower models \cite{wang2019two-tower}. They suffer from the above issue due to a lack of interactions between the two modalities. Other works take one modality as auxiliary knowledge to ground the other, such as
\emph{(i)} exploiting the textual representation to augment a graph reasoning model \cite{lv2020PLM2KG2}, and \emph{(ii)} augmenting the language representation of a QA example with the encoded graph knowledge \cite{kagneg19,yang2019KG2PLM3}.
As for these methods, the information flows one way between the two modalities, which still limits the knowledge interaction. 

More recently, GreaseLM mixes the multi-modal representations in the intermediate layers of the LM and the GNN to enable two-way interactions between both modalities. However, GreaseLM neglects to adapt the primary representations of PLMs and KGs to each other, thus poorly seeding the joint structure. In order to bridge the gaps between PLMs and KGs before joint learning, we propose a knowledge adaptive post-training objective, which brings a better starting point to the joint reasoning module. Simultaneously, we retain the structure of GreaseLM as our backbone model.

\begin{figure*}[t]
\centering
\includegraphics[width=\textwidth]{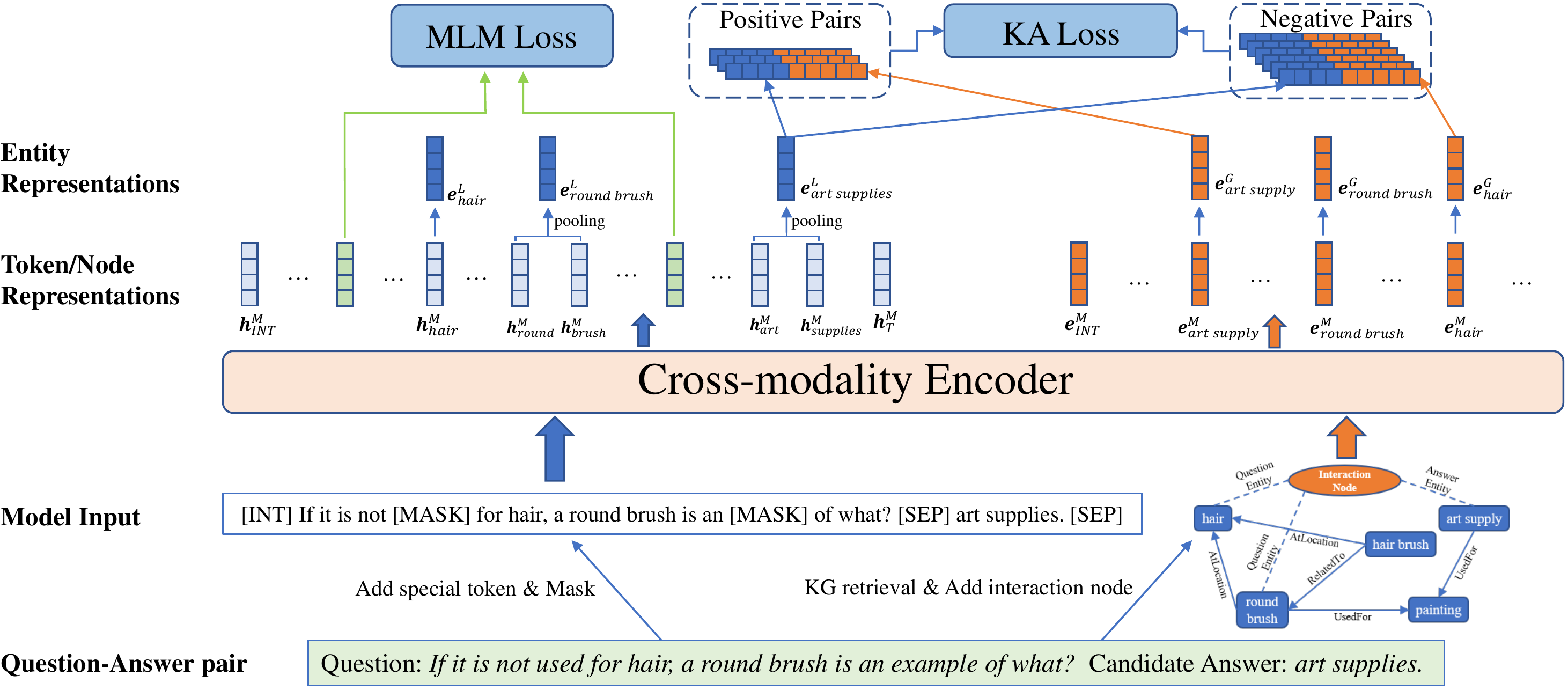} 
\caption{An overview of the post-training process. First, we transform the question-answer pair to model input (\emph{i.e.}, masked context and KG sub-graph), and use the cross-modality encoder to get the fused token(node) representations $\textbf{h}^M$($\textbf{e}^M$). Then \emph{(i)} the token representations corresponding to the [MASK] token are used to calculate the MLM loss, and \emph{(ii)} the entity representations $\textbf{e}^L$ for text entities and $\textbf{e}^G$ for KG nodes are used to calculate the knowledge adaptive (KA) loss. }
\label{Fig.post-training}
\end{figure*}

Additionally, due to the imperfectness of the KG retrieval module, models have to reason with inadequate and noisy knowledge retrieved from KGs. Although some works evaluate KAQA model robustness in severe settings, how to explicitly improve this ability remains an open question \cite{gu2021beyond}. To this end, in addition to the supervision signal, we design two auxiliary self-supervised fine-tuning objectives to improve model performance in utilizing useful knowledge and distinguishing irrelevant knowledge.

\section{Proposed Method}
\label{proposed method}
In this section, we will explain the problem of KAQA we are investigating and introduce the objectives for our post-training and fine-tuning methods. The pipeline of our two-stage training framework is illustrated in Figure \ref{Fig.pipeline}.

\subsection{Problem Formalization}
\label{problem pormulation}
We mainly focus on the knowledge-aware multiple choice question answering (KAMCQA) task. Generally speaking, an MCQA-type dataset consists of examples with a context paragraph $c$, a question $q$, and a candidate answer set $A$, all in text format. Treating $c$ and $q$ as a whole, an MCQA example can be regarded as a Q-A pair. The KAMCQA task is an extension of MCQA, where an external knowledge graph $G$ is accessible to provide auxiliary knowledge relevant to a given MCQA example.

In practice, due to computational factors, a KAQA system first retrieves a sub-graph $G_{sub}$ from $G$ for each MCQA example which consists of a certain number of entities that are most relevant to that example. We follow the procedure from \citet{QA-GNN} to compute the relevance score between each KG entity and the MCQA example.
Given an example $(c, q, A)$ and the retrieved $G_{sub}$ as input, the task is to identify which answer $a \in A$ is correct.
For simplicity, when computing the probability of a candidate answer being the correct answer, we refer to that answer as $a$.

\subsection{Cross-modality Encoder}
The cross-modality encoder extracts and fuses the information from text and KG.
We use GreaseLM as the cross-modality encoder and give a brief introduction here (see more details in our supplementary material). GreaseLM employs a PLM to encode the textual input $(c, q, a)$ and a GNN model to process the $G_{sub}$, and further use a two-layer MLP to mix these representations.
Specifically, the textual context is appended with a special $\textbf{interaction token}$ and passed through $N$ LM-based unimodal encoding layers to get the pre-encoded language representations $\{\textbf{h}_{int}, \textbf{h}_1,...,\textbf{h}_T\}$. Simultaneously, an $\textbf{interaction node}$, whose representation will be used to interact with the representation of the interaction token, is also added to the $G_{sub}$. In each of the following $M$ GreaseLM layers, the language representations are fed into transformer LM encoder blocks that continue to encode textual context, and graph representations are fed into a GNN layer to perform a round of information propagation
between nodes in the graph:
\begin{gather}
    \{\Tilde{\textbf{h}}_{int}^l, \textbf{h}_1^l, ..,\textbf{h}_T^l\} = \operatorname{LM-Enc}\left(\{\textbf{h}_{int}^{l\!-\!1}, \textbf{h}_1^{l\!-\!1},\! ..,\textbf{h}_T^{l\!-\!1}\!\}\right)\! \\
    \{\Tilde{\textbf{e}}_{int}^l,\textbf{e}_1^l,\! ..., \textbf{e}_J^l\} = \operatorname{GNN}\left(\{\textbf{e}_{int}^{l-1}, \textbf{e}_1^{l-1}\!,...,\textbf{e}_J^{l-1}\!\}\right)
\end{gather}
Then the representations of the interaction token and the interaction node are concatenated and passed through the MLP to get a modality-wise mixed representation:
\begin{equation}
    \left[\textbf{h}_{int}^l; \textbf{e}_{int}^l\right] = \operatorname{MLP}\left(\left[\Tilde{\textbf{h}}_{int}^l; \Tilde{\textbf{e}}_{int}^l\right]\right)
\end{equation}
Consequently, the cross-modality encoder ensures that information propagates between both modalities.
\begin{figure*}[t]
\centering
\includegraphics[width=\textwidth]{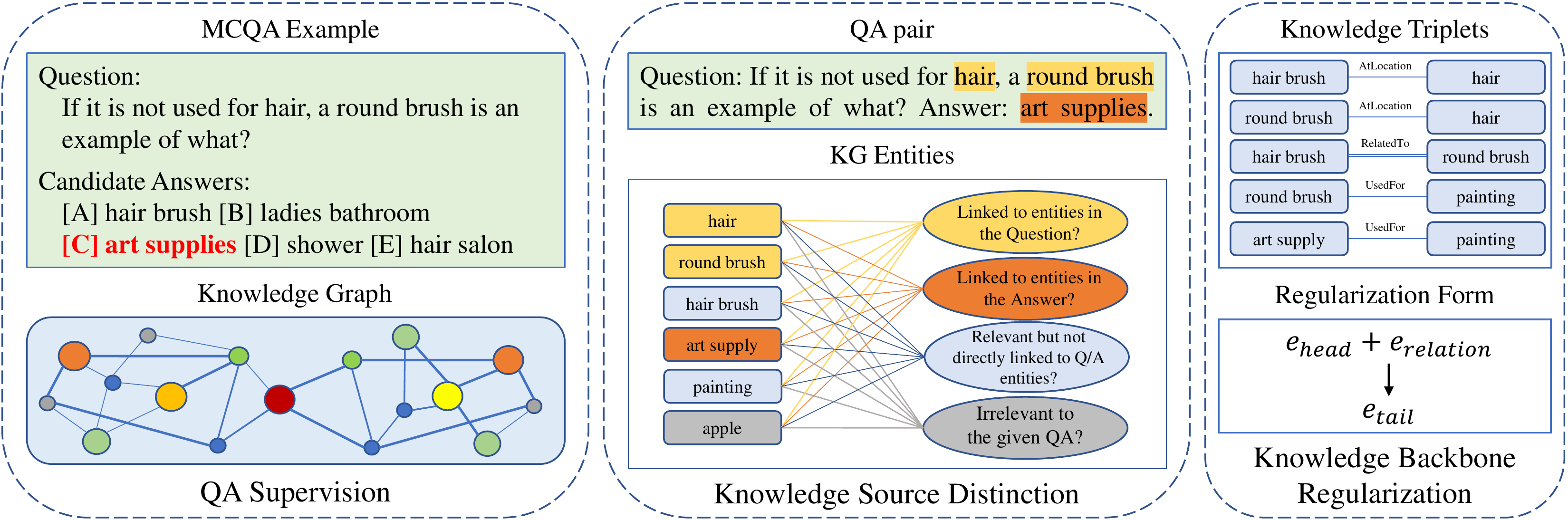} 
\caption{An overview of the knowledge-aware fine-tuning objectives.}
\label{Fig.multi-task}
\end{figure*}
\subsection{Knowledge Adaptive Post-training}
\label{post-training}
In fusing the representations from PLMs and KGs, two main challenges exist: \emph{(i)} since PLMs and KGs are usually pre-trained on different corpora, the representations from the two modalities may be semantically contradictory, thus leading to confusion in joint reasoning; and \emph{(ii)} the distribution of them may be discordant, which means that even similar semantics may be embedded in an opposite direction in the latent space.
In order to close the gap between both modalities, \emph{i.e.}, provide a better initialization for fine-tuning, we propose a knowledge adaptive (KA) loss in addition to the MLM \cite{devlin2018bert} loss to perform post-training.

As shown in Figure \ref{Fig.post-training},
we get token representations $\textbf{H}^M=\{\textbf{h}_{int}^M, \textbf{h}_1^M, ..., \textbf{h}_T^M\}$ and KG node representations $\textbf{E}^M=\{\textbf{e}_{int}^M, \textbf{e}_1^M, ...,\textbf{e}_K^M\}$ through the cross-modality encoder. We then obtain text entity representations $\textbf{E}^L=\{\textbf{e}_{int}^L, \textbf{e}_1^L, ...,\textbf{e}_J^L\}$ by pooling over the corresponding token representations, where each $\textbf{e}_i^L \in \textbf{E}^L$ represents an entity in the input Q-A pair. Meanwhile, we regard KG node representations as KG entity representations $\textbf{E}^G=\{\textbf{e}_{int}^G, \textbf{e}_1^G, ...,\textbf{e}_J^G\}$, \emph{i.e.}, $\textbf{E}^G=\textbf{E}^M$, because each node in the KG represents an entity.
Then, we adopt a contrastive learning framework to align the inherent knowledge in $\textbf{E}^L$ and $\textbf{E}^G$, where each pair of $\textbf{e}^L_i$ and $\textbf{e}^G_j$ that represents the same entity constitutes a positive pair (\emph{e.g.}, $\textbf{e}_{art\ supplies}^{L}$ and $\textbf{e}_{art\ supply}^G$ in Figure \ref{Fig.post-training}) and $\textbf{e}^L_i$ is treated as a negative example for entity representations other than $\textbf{e}^G_j$ in $\textbf{E}^G$, vice versa.
Given a positive or negative pair of entity representations, we concatenate them and pass the joint representation through an MLP to get the probability $\hat{y} \in R$ that indicates whether the two entities are matched:


\begin{equation}
    \hat{y} = \textbf{W}_1\operatorname{ReLU} \left(\textbf{W}_0\left[\textbf{e}_i^{L}; \textbf{e}_j^{G}\right] + \textbf{b}_0\right)
\end{equation}
where $\textbf{W}_1 \in R^{d}$, $\textbf{W}_0 \in R^{d\times d}$ and $\textbf{b}_0 \in R^{d}$ are trainable parameters, $d = d_{l} + d_{g}$, $d_l$ and $d_g$ are the dimension of $\textbf{e}_i^{L}$ and $\textbf{e}_j^{G}$, respectively.

For each Q-A pair, we choose k positive pairs and k corresponding negative pairs. We use labels $\textbf{y} = \left[y_1, y_2, ..., y_{2k}\right]$ to 
distinguish positive and negative pairs, where $y_i = 1$ if and only if the $i$-th one is a positive pair, otherwise $y_i = 0$. The calculation of the KA loss is as follows:
\begin{equation}
    \mathcal{L}_{KA} = - \frac{1}{2k}\sum_{i=1}^{2k} \left(1-y_i\right)\log\left(1-\hat y_{i}\right) 
    + y_i\log\left(\hat y_{i}\right)
\end{equation}

Similar to BERT \cite{devlin2018bert}, we randomly choose $15\%$ tokens in the MCQA example to perform MLM. The overall post-training loss is the sum of the two losses:
\begin{equation}
    \mathcal{L}_{post} = \mathcal{L}_{KA} + \mathcal{L}_{MLM}
\end{equation}

\subsection{Knowledge-aware Fine-tuning}
\label{multi-task finetuning}
To fully exploit the encoded knowledge in PLMs and KGs, we propose two auxiliary self-supervised learning objectives in addition to the QA supervision signal.

\subsubsection{QA Supervision}
We follow the procedure from \citet{GreaseLM} to formulate the fully-supervised objective: for a $n$-way MCQA example $(c, q, A)$, the probability $p\left(a_i | q, c\right)$ that $a_i \in A$ is the correct one is computed as:
\begin{equation}
    p\left(a_i\ |\ q, c\right) \propto \operatorname{exp}\left(\operatorname{MLP}\left(\textbf{h}_{int}^{M}, \textbf{e}_{int}^{M}, \textbf{g}\right)\right)
\end{equation}
where $\textbf{g}$ is computed by attentively pooling over the KG node embeddings $\{\textbf{e}_1^M,...,\textbf{e}_J^M\}$ using $\textbf{h}_{int}^{M}$
as query. 
Then we compute the cross-entropy loss:


\begin{equation}
    \mathcal{L}_{Sup}=-\sum_{i=1}^ny_i\log\left(p\left(a_i \ |\ q, c\right)\right)
\end{equation}
where $y_i = 1$ if $a_i$ is the correct answer, otherwise $y_i = 0$.

In the inference time, the answer is predicted by:
\begin{equation}
a_p = \operatorname{argmax}_{a \in A}\ p\left(a\ |\  q, c\right)    
\end{equation}

\subsubsection{Knowledge Source Distinction}
\label{knowledge source distinction}
In the process of KG retrieval, entities related to the question or answer and edges connecting these entities are retrieved to form a sub-graph. Distinguishing whether a retrieved entity is related to the question or the answer is an essential aspect of knowledge understanding.
For example, 

\begin{center}
\emph{A weasel has a thin body and short legs to easier burrow after prey in a what?} \\
\emph{(A) tree (B) mulberry bush (C) chicken coop} \\
\emph{(D) viking ship (E) rabbit warren}    
\end{center}

\noindent To pick out the correct answer \emph{(E) rabbit warren}, knowledge about the predator-prey relationship and the narrowness of the prey's lair is required. Under real circumstances, preparing for this knowledge will introduce noise like weasels' dens are narrow. So the model has to be clear that rabbits, in the candidate answer (E), have narrow warrens, while weasel appears in the question as the predator. Otherwise, guided by the ambiguous noisy knowledge, the wrong candidate answer (C) may be chosen.

Moreover, due to the imperfectness of the KG retrieval module, suspicious entities are often introduced to $G_{sub}$, \emph{i.e.}, containing lots of irrelevant entities. Thus, it is important to let the model distinguish whether a retrieved entity is relevant to the Q-A pair.
However, distinguishing whether a retrieved entity is relevant to the context is quite challenging due to lack of supervision, so we want to train the model in a heuristic way.
To this end, we manually add $k_{irr}$ irrelevant entities to $G_{sub}$
to guide the model to learn distinguishable representations for significantly irrelevant entities and gradually differentiate existing irrelevant entities from the others.

To achieve the above two goals, given the representations $\textbf{E}^G=(\textbf{e}^G_1,\textbf{e}^G_2,\dots,\textbf{e}^G_{m+k_{irr}})$ of $m+k_{irr}$ entities in $G_{sub}$, we formulate the knowledge source distinction loss as:

\begin{equation}
\begin{split}
    \mathcal{L}_{KSD}=-\sum_{i=1}^{m+k_{irr}}\sum_{j=1}^4y_{ij}\log\left(\hat y_{ij}\right)
\end{split}
\end{equation}

\begin{equation}
    \hat{\textbf{y}}_i = \operatorname{softmax}\left(\textbf{W}_3\operatorname{ReLU} \left(\textbf{W}_2\textbf{e}_i^G + \textbf{b}_1\right)\right)
\end{equation}
where $\textbf{W}_3 \in R^{4\times d_g}$, $\textbf{W}_2 \in R^{d_g\times d_g}$ and $\textbf{b}_1 \in R^{d_g}$ are trainable parameters, $d_g$ is the dimension of $\textbf{e}_i^G$, $\textbf{y}_i=[y_{i1},\dots,y_{i4}]$ is the one-hot vector indicating whether the $i$-th KG entity is: (1) linked to entities in the question, (2) linked to entities in the candidate answer, (3) an entity in the retrieved multi-hop neighborhood but not directly linked to mentioned entities, or (4) an irrelevant entity.

\subsubsection{Knowledge Backbone Regularization}
\label{knowledge backbone regularization}
The knowledge backbone regularization objective is designed to guide the model to better understand the internal relation of KG knowledge triplets $<h, r, t>$, where $h$ and $t$ are head entity and tail entity, respectively, $r$ is the relationship between them. Inspired by TransE \cite{bordes2013regularization}, where relationships are represented as translations in the embedding space, we assume that, in the latent representation space, the summation of the head entity representation $\textbf{e}_h$ and the relationship representation $\textbf{e}_r$ should be close to the tail entity representation $\textbf{e}_t$ as much as possible, \emph{i.e.}, $\textbf{e}_h + \textbf{e}_r \rightarrow \textbf{e}_t$.

To this end, given the entity and relationship representations produced by the cross-modality encoder, we introduce a regularization for the $k_{reg}$ knowledge triplets:

\begin{equation}
    \mathcal{L}_{KBR} = \sum_{i=1}^{k_{reg}}\left(1-\operatorname{cos}\left(\textbf{e}_{h_i}+\textbf{e}_{r_i}, \textbf{e}_{t_i}\right)\right)
\end{equation}

The overall fine-tuning loss is as follows:

\begin{equation}
    \mathcal{L}_{finetune} = \mathcal{L}_{Sup} + \mathcal{L}_{KSD} + \mathcal{L}_{KBR}
\end{equation}

\section{Experimental Setups}
See implementation details in our supplementary material.
\subsection{Datasets}
We evaluate our proposed post-training and fine-tuning methods on three MCQA datasets: CommonsenseQA \cite{commonsenseQA} and OpenbookQA \cite{openbookQA} for commonsense reasoning, and MedQA-USMILE \cite{jin202medqa} as a medical QA benchmark.

The \textbf{CommonsenseQA} dataset includes 12,102 5-way multiple-choice questions. Each question requires extra commonsense knowledge beyond the surface-level textual information given by the context. Due to the fact that the official test is hidden, our experiments are conducted using the in-house data split of \citet{kagneg19}.

The \textbf{OpenbookQA} dataset consists of 5,957 4-way multiple-choice questions about elementary scientific knowledge, along with an open book of scientific facts. We perform experiments using the official data splits of \citet{obqaSplit}.

The \textbf{MedQA-USMILE} dataset includes 12,723 4-way multiple-choice questions that are originally collected from the National Medical Board Examination in the USA. These questions assess a model's ability to apply medical and clinical knowledge, concepts, and principles.

\subsection{Baselines}

For commonsense reasoning tasks, we include Roberta-Large \cite{roberta} and several advanced knowledge graph enhanced question answering systems as baselines for comparison: (1) RGCN \cite{RGCN}, (2) KagNet \cite{kagneg19}, (3) MHGRN \cite{MHGRN}, (4) QA-GNN \cite{QA-GNN}, and (5) GreaseLM \cite{GreaseLM}. All of these methods, except for Roberta-Large, follow the paradigm of LM+KG. GreaseLM is the top-performing one that fuses representations from the LM and KG with modality interaction layers. 
For MedQA-USMILE, besides LM+KG methods, we compare with BioBERT-Large \cite{lee2020biobert}, a pre-trained biomedical language representation model based on BERT-Large \cite{devlin2018bert}, and SapBERT \cite{liu2021sapbert}, a state-of-the-art model for medical entity representation learning, which improves model's ability to capture entity relationships with the help of entity disambiguation objectives.







\begin{table}[t]
\small
\centering
\setlength{\tabcolsep}{14pt}
\begin{tabular}{lc}
\toprule
\textbf{Model} & \textbf{IHtest-Acc} (\%)\\
\hline
\noalign{\smallskip}
RoBERTa-Large (w/o KG)$^{\clubsuit}$ &  68.7 \\
RGCN \cite{RGCN}$^{\clubsuit}$  & 68.4 \\
KagNet \cite{kagneg19}$^{\clubsuit}$  & 69.0 \\
MHGRN \cite{MHGRN}$^{\clubsuit}$  & 71.1  \\
QA-GNN \cite{QA-GNN}$^{\clubsuit}$  & 73.4  \\
GreaseLM \cite{GreaseLM}$^{\clubsuit}$  & 74.2  \\
\hline
\noalign{\smallskip}
GreaseLM (Our implementation)  & 73.6  \\
+ \textbf{FiTs} (\textbf{Ours})  & \textbf{76.2}  \\
\bottomrule
\end{tabular}
\caption{Performance comparison on CommonsenseQA in-house split. We report the in-house Test (IHtest) accuracy using the data split of \citet{kagneg19}, because the official test is hidden. ${\clubsuit}$: results from \citet{GreaseLM}; all other results are reproduced by ourselves.
}
\label{csqa result}
\end{table}




\begin{table}[t]
\small
\centering
\setlength{\tabcolsep}{16pt}
\begin{tabular}{lc}
\toprule
\textbf{Model} & \textbf{Test-Acc} (\%) \\
\hline
\noalign{\smallskip}
AristoRoBERTa (no KG)$^{\clubsuit}$ & 78.4 \\
RGCN \cite{RGCN}$^{\clubsuit}$  & 74.6 \\
MHGRN \cite{MHGRN}$^{\clubsuit}$ & 80.6  \\
QA-GNN \cite{QA-GNN}$^{\clubsuit}$ & 82.8  \\
GreaseLM \cite{GreaseLM}$^{\clubsuit}$ & 84.8  \\
\hline
\noalign{\smallskip}
GreaseLM (Our implementation) & 84.2 \\
+ \textbf{FiTs} (\textbf{Ours}) & \textbf{86.0} \\
\bottomrule
\end{tabular}
\caption{Test accuracy comparison on OpenBookQA. $\clubsuit$: results from \citet{GreaseLM}; all other results are reproduced by ourselves.}
\label{obqa result}
\end{table}

\begin{figure*}[t]
\centering
\includegraphics[width=\textwidth]{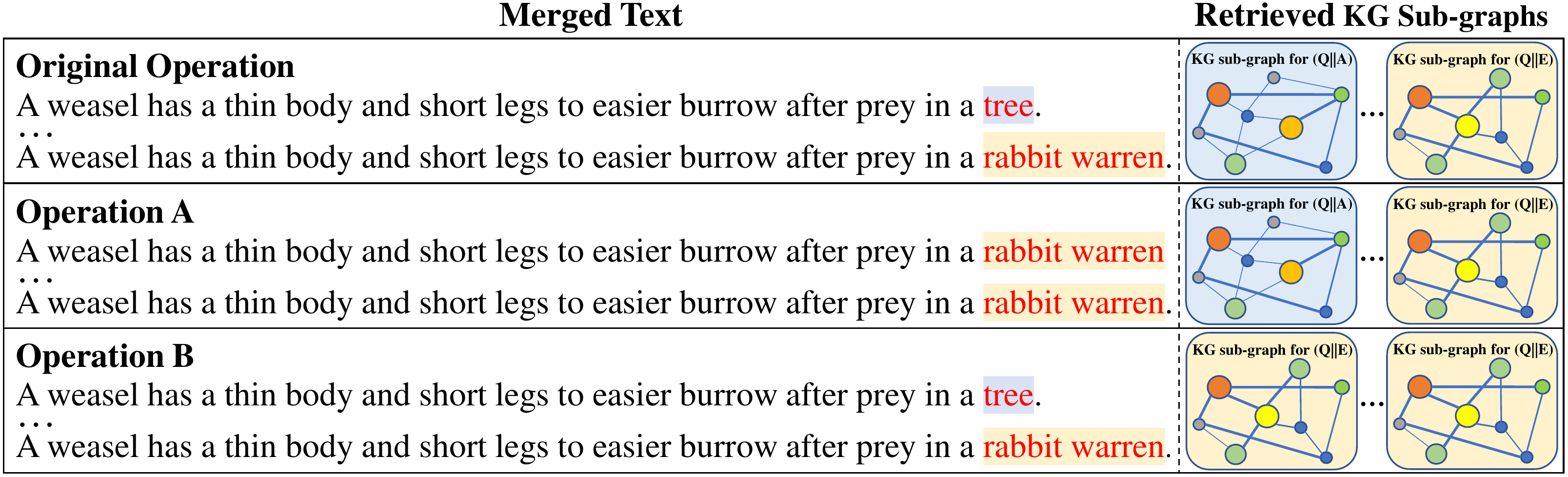} 
\caption{Operations A and B are used for quantitative analysis (section \ref{sec:quantitative analysis}). The unprocessed MCQA example is shown in Figure \ref{Fig.example}. $Q || A$ denotes the text obtained by merging the question $Q$ and the candidate answer $A$.}
\label{Fig.quantitative analysis}
\end{figure*}

\section{Results and Analysis}
Our results in Tables \ref{csqa result} and \ref{obqa result} demonstrate consistent improvements on the CommonsenseQA (CSQA) and OpenBookQA (OBQA) datasets. On CSQA, our model's test performance improves by 7.5\% over RoBERTa-Large (fine-tuned LM without KG), 2.8\% over LM+KG methods before GreaseLM, and 2.6\% over our implementation of the best prior LM+KG system, GreaseLM, on which we evaluate our proposed post-training and fine-tuning methods. As for the OBQA dataset, our model's test performance improves by 7.6\% over AristoRoBERTa (finetuned LM without KG), 3.2\% over existing LM+KG systems other than GreaseLM, and 1.8\% over the GreaseLM that we have implemented. The boost over GreaseLM reveals the superiority of our proposed knowledge-adaptive post-training and knowledge-aware fine-tuning methods in making use of the inherent knowledge from PLMs and KGs.


\begin{figure*}[t]
\centering
\includegraphics[width=\textwidth]{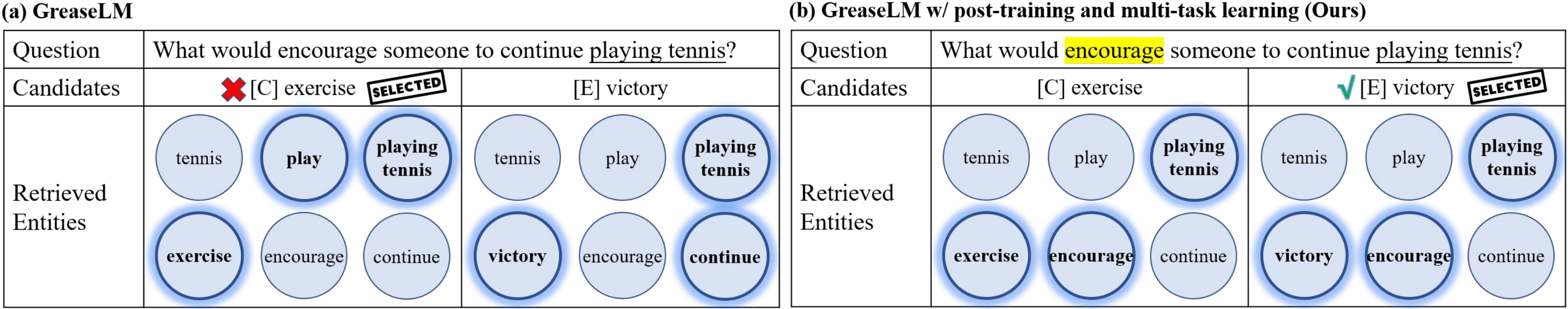} 
\caption{Attention analysis of GreaseLM w/ and w/o our methods. Entities with higher attention weights are highlighted. Our model demonstrates the expected pattern by consistently focusing on the ``\texttt{encourage}" entity. }
\label{Fig.qualitative analysis}
\end{figure*}

\subsection{Quantitative Analysis}
\label{sec:quantitative analysis}
Empirically, both the parametric knowledge and the joint reasoning ability of the model benefit KAQA tasks \cite{longpre2021quantitiveAnalysis}. Provided the overall performance improvements, we investigated whether both knowledge sources make contributions or not by conducting new operations on the dataset. Originally, given an unprocessed MCQA example (where a question and candidate answers are separate), we separately merge the question and each candidate answer together and obtain KG sub-graphs based on the merged text to generate training/validation/testing data (the Original Operation in Figure \ref{Fig.quantitative analysis}). To evaluate the contribution of the model's joint reasoning ability, we apply \textbf{operation A}, where each candidate answer is replaced with the correct answer while their retrieved sub-graphs remain unchanged. This operation restricts the model to infer only based on the question and the KG sub-graph. As for the model's parametric knowledge, we apply \textbf{operation B}, where each sub-graph is replaced with the one obtained based on the correct answer while all candidate answers remain unchanged. 
The KG knowledge provided is equivalent and relevant to the correct answer, so the model is forced to make judgments based on an understanding of the relationship between a question and its candidate answers.

We conduct experiments using the IHtest set of CommonsenseQA. The new test set obtained based on operation A is named test-reasoning, test-reason for short, and that based on operation B is named test-parametric, test-param for short. The results in Table \ref{csqa-new-test-test} demonstrate that both the proposed post-training and fine-tuning methods can significantly increase the model's parametric knowledge. In comparison, improvements in the model's joint reasoning ability are marginal, which may be partly due to the imperfectness of the KG retrieval module, \emph{i.e.}, a bottleneck for reasoning.

\begin{table}[t]
\small
\centering
\setlength{\tabcolsep}{5pt}
\begin{tabular}{lcc}
\toprule
\textbf{Model} & \textbf{test-reason} & \textbf{test-param}\\
\hline
\noalign{\smallskip}
GreaseLM \cite{GreaseLM}  & 73.4 & 69.0 \\
+ post-training & 73.9 & 70.9 \\
+ knowledge-aware fine-tuning &74.2 & 71.2\\
+ \textbf{FiTs} (\textbf{Ours}) & \textbf{74.5} & \textbf{71.6} \\
\bottomrule
\end{tabular}
\caption{The test-reason set evaluates models' joint reasoning ability, while the test-param set measures models' parametric knowledge.}
\label{csqa-new-test-test}
\end{table}

\subsection{Qualitative Analysis}
Analyzing the attention mechanism is a crucial way to examine a model's behavior.
In this section, we analyze graph attention weights in the last layer of GAT (\emph{i.e.}, the attention weight between each entity in $G_{sub}$ and the context) to find out which entity the model focuses on and investigate whether our model demonstrates a sensible reasoning process.
As the example in Figure \ref{Fig.qualitative analysis} demonstrates, while GreaseLM concentrates on ``\texttt{play}" (given the candidate answer [C] exercise) and ``\texttt{continue}" (given the candidate answer [E] victory), our model shows consistent interest on ``\texttt{encourage}", which is the core factor in clarifying the relationship between a candidate answer and ``\texttt{playing tennis}". Intuitively, our model performs a more expected behavior of human reasoning, suggesting the success of our fine-grained two-stage training framework.

\begin{table}[t]
\small
\centering
\setlength{\tabcolsep}{16pt}
\begin{tabular}{lc}
\toprule
\textbf{Model} & \textbf{Test-Acc} (\%) \\
\hline
\noalign{\smallskip}
BioBERT-Base \cite{lee2020biobert}$^\clubsuit$ & 34.1  \\
BioBERT-Large \cite{lee2020biobert}$^\clubsuit$ & 36.7 \\
QA-GNN \cite{QA-GNN}$^\clubsuit$ & 38.0 \\
GreaseLM \cite{GreaseLM}$^\clubsuit$ & 38.5 \\
\hline
\noalign{\smallskip}
GreaseLM (Our implementation) & 38.6 \\
+ \textbf{FiTs} (\textbf{Ours}) & \textbf{39.2} \\
\bottomrule
\end{tabular}
\caption{Test accuracy comparison on MedQA-USMLE. $\clubsuit$: results from \citet{GreaseLM}; all other results are reproduced by ourselves.}
\label{medqa result}
\end{table}

\subsection{Domain Generality}
So far, our model's performance demonstrates the effectiveness of our proposed post-training and fine-tuning methods in the commonsense reasoning domain. Here we further investigate whether they are applicable in the medical domain. Following \citet{GreaseLM}, we evaluate our model's performance on the MedQA-USMLE dataset. The results in Table \ref{medqa result} illustrate the generality of our methods, achieving an improvement of 0.6\% over the backbone model.

\subsection{Ablation Studies}
We investigated the impact of each part of the post-training and fine-tuning methods through a series of ablation experiments using the ConmensenseQA IHtest set.

\begin{figure}[t]
\centering
\includegraphics[width=0.47\textwidth]{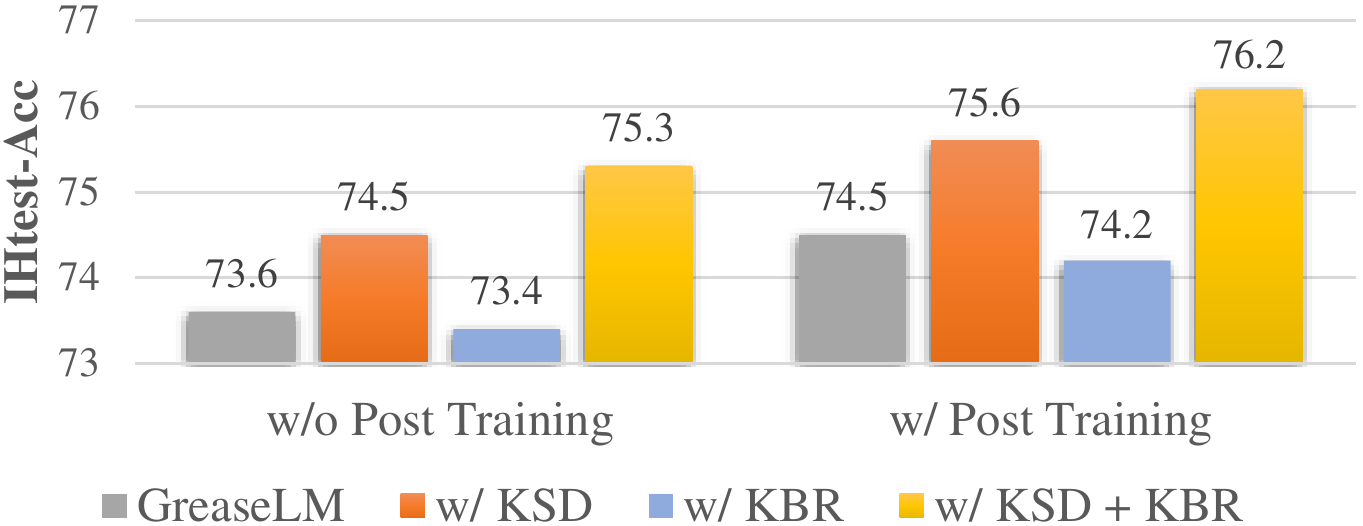} 
\caption{Ablation studies of the fine-tuning objectives.}
\label{Fig.ablation}
\end{figure}

\noindent \textbf{Post-training:} Results in Table \ref{ablation-post} suggest that both MLM and KA are helpful for improving the model's test performance, and they complement each other, bringing an improvement by 0.9\% over raw GreaseLM. Additionally, the comparison between the left and right half of Figure \ref{Fig.ablation} demonstrates that our post-training method clearly benefits the subsequent fine-tuning process. 

\noindent \textbf{Fine-tuning:} The usefulness of the two self-supervised objectives and the interdependence between them are demonstrated in Figure \ref{Fig.ablation}. Specifically, with or without post-training, \emph{(i)} KSD alone can significantly improve model performance, but KBR alone degrades that because the model without KSD lacks discrimination on the relevance and importance of the external knowledge; \emph{(ii)} KSD and KBR together bring the best result, suggesting that these two objectives complement each other—KSD improves model's discrimination on the relevance and importance of the external knowledge; KBR acts as a commonsense-knowledge-oriented regularization to avoid task-specific overfitting.

\begin{table}[t]
\small
\centering
\setlength{\tabcolsep}{18pt}
\begin{tabular}{lc}
\toprule
\textbf{Model} & \textbf{IHtest-Acc}\\
\hline
\noalign{\smallskip}
GreaseLM (Only QA supervision) & 73.6 \\
+ MLM & 74.1 \\
+ KA & 74.3  \\
+ MLM + KA & \textbf{74.5}  \\
\bottomrule
\end{tabular}
\caption{Ablation studies of the post-training objectives.}
\label{ablation-post}
\end{table}

\section{Conclusion}
This paper introduces \textbf{FiTs}, a fine-grained two-stage training framework for KAQA tasks, including the knowledge adaptive post-training stage (with MLM and KA objectives) and the knowledge-aware fine-tuning stage (with KBR and KSD objectives). 
Post-training alleviates the gaps between the representations from PLMs and KGs, leading to a better starting point for fine-tuning. 
Fine-tuned with the proposed objectives, the model is better at identifying how relevant and essential each entity in the retrieved KG sub-graph is to the given question. Experimental results on benchmarks in the commonsense reasoning and medical domains show the great improvements our method brings to the backbone model, which are further demonstrated to be reflected in both model's parametric knowledge and joint reasoning ability. Our work reveals a better way to integrate knowledge from PLMs and KGs and gives insights on how to design learning objectives for KAQA tasks.

\section{Acknowledgement}
This paper was partially supported by Shenzhen Science \& Technology Research Program (No: GXWD202012311658-\\07007-20200814115301001) and NSFC (No: 62176008)

\bibliography{aaai23}

\end{document}